%% file: main.tex
\documentclass[10pt,twocolumn,letterpaper]{article}

\usepackage{iccv}              

\usepackage{tikz}
\usepackage{amsmath}
\usetikzlibrary{positioning}
\usepackage{pgfplots}
\pgfplotsset{compat=1.18}
\usepackage{placeins}
\usepackage{stackengine}
\usepackage{multirow}
\usepackage[table,xcdraw]{xcolor}
\usepackage{booktabs}
\usepackage{algorithm}
\usepackage{algorithmic}
\usepackage{comment}
\usepackage{bm}
\usetikzlibrary{shapes.geometric, arrows.meta, positioning}


\definecolor{iccvblue}{rgb}{0.21,0.49,0.74}
\usepackage[pagebackref,breaklinks,colorlinks,allcolors=iccvblue]{hyperref}

\def\paperID{} 
\def\confName{}
\def\confYear{}

\title{Sequence-Adaptive Video Prediction in Continuous Streams using Diffusion Noise Optimization}

\author{
Sina Mokhtarzadeh Azar\textsuperscript{1,4} \quad
Emad Bahrami\textsuperscript{1,4} \quad
Enrico Pallotta\textsuperscript{1,4} \quad
Gianpiero Francesca\textsuperscript{2} \\[0.1cm]
Radu Timofte\textsuperscript{3} \quad
Juergen Gall\textsuperscript{1,4} \\[0.3cm]
\textsuperscript{1}University of Bonn \quad
\textsuperscript{2}Toyota Motor Europe \quad
\textsuperscript{3}University of Wuerzburg \\
\textsuperscript{4}Lamarr Institute for Machine Learning and Artificial Intelligence 
}

\begin{document}

\twocolumn[{%
\renewcommand\twocolumn[1][]{#1}%
\maketitle

}]

\newcommand{\myparagraph}[1]{\vspace{0.5em}\noindent\textbf{#1}}
\input{sec/0_abstract}

\input{sec/1_intro}

\input{sec/2_relatedwork}
\input{sec/3_method}
\input{sec/4_experiments}

\input{sec/5_conclusion}
{
    \small
    \bibliographystyle{ieeenat_fullname}
    \bibliography{main}
}

\input{sec/X_suppl}

\end{document}

%% file: sec/0_abstract.tex
\begin{abstract}
In this work, we investigate diffusion-based video prediction models, which forecast future video frames, for continuous video streams. In this context, the models observe continuously new training samples, and we aim to leverage this to improve their predictions. We thus propose an approach that continuously adapts a pre-trained diffusion model to a video stream. Since fine-tuning the parameters of a large diffusion model is too expensive, we refine the diffusion noise during inference while keeping the model parameters frozen, allowing the model to adaptively determine suitable sampling noise. We term the approach Sequence Adaptive Video Prediction with Diffusion Noise Optimization (SAVi-DNO). 
To validate our approach, we introduce a new evaluation setting on the Ego4D dataset, focusing on simultaneous adaptation and evaluation on long continuous videos. Empirical results demonstrate improved performance based on FVD, SSIM, and PSNR metrics on long videos of Ego4D and OpenDV-YouTube, as well as videos of UCF-101 and SkyTimelapse, showcasing SAVi-DNO's effectiveness.
\end{abstract}

%% file: sec/1_intro.tex
\begin{figure}[t]
\centering
   \includegraphics[trim=60 450 60 0, width=1.0\linewidth]{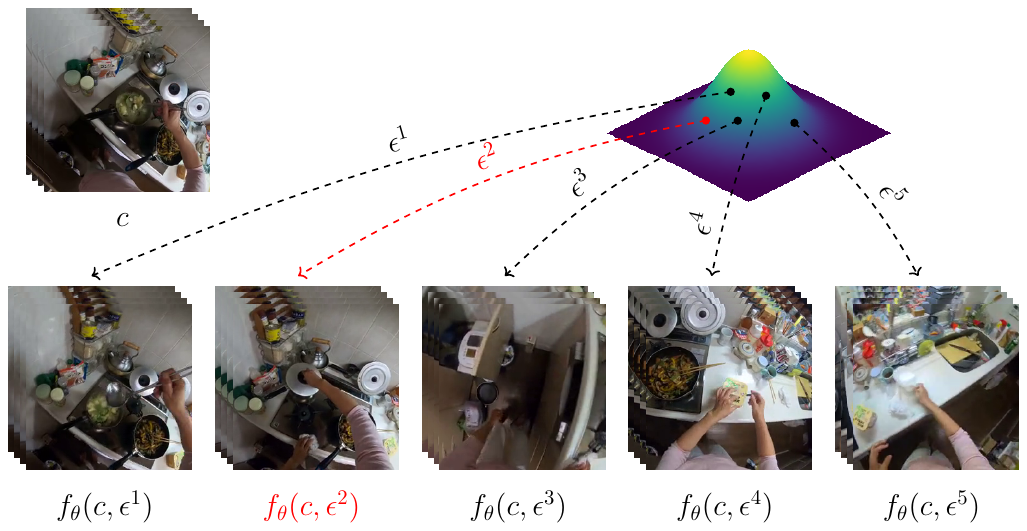}
   \caption{
   Given an observation from the past $c$, a diffusion model for video prediction can create multiple predictions by sampling noise $\epsilon$ from a Gaussian distribution. If the continuous video stream changes its distribution compared to the training data, the correct future frames might have a very low probability to be sampled. In this work, we therefore propose to change the way how $\epsilon$ is sampled in continuous video streams.
   }
   \label{fig:teaser}
\end{figure}

\section{Introduction}
\label{sec:intro}
Predicting future video frames is useful in multiple applications such as autonomous driving, robotics, human-computer interaction, and augmented reality. In recent years, approaches based on diffusion models have shown very good results \cite{ho2020denoising,hoppe2022diffusion,voleti2022mcvd,lu2024vdt, zhang2024extdm}. Current approaches, however, separate the training from the inference. They are first trained on a large dataset and then applied to video clips for predicting future video frames. In the context of continuous video streams, which is the most relevant scenario, the models observe continuously new training samples since after each prediction, the observation of the prediction arrives. While storing the data is often prohibited due to legal and ethical reasons, the data can be used to adapt the diffusion model continuously to the video streams. 

While updating the parameters of a pre-trained diffusion model is an option, it is very inefficient and causes concerns that private data will be stored in the model. In this work, we thus propose an alternative to adapt diffusion models for video prediction to continuous video streams. As illustrated in Fig.~\ref{fig:teaser}, diffusion models predict future video frames from the past frames by sampling from a Gaussian distribution. For a given observation, the prediction can thus be changed by either updating the parameters of the model or changing the noise. The latter has the advantage that it is much faster to update and that the model does not store any private data, which is a very important aspect for privacy-sensitive applications. To this end, we continuously optimize the sampling noise when new observations for a prediction arrive and add some additional noise. 

We evaluate the approach on four datasets for two different diffusion models, namely PVDM \cite{yu2023video} and Vista \cite{gao2025vista}. The datasets include the SkyTimelapse~\cite{xiong2018learning} and UCF-101~\cite{soomro2012ucf101} datasets, which have relatively short videos, the Ego4D \cite{grauman2022ego4d} dataset, which contains long and challenging egocentric videos, and the OpenDV-YouTube \cite{yang2024generalized} dataset, which also contains long videos. We demonstrate that our approach improves FVD, SSIM, and PSNR for all datasets and diffusion models.

%% file: sec/2_relatedwork.tex
\begin{figure*}[t]
    \centering
   \includegraphics[trim=0 0 0 0, width=0.8\linewidth]{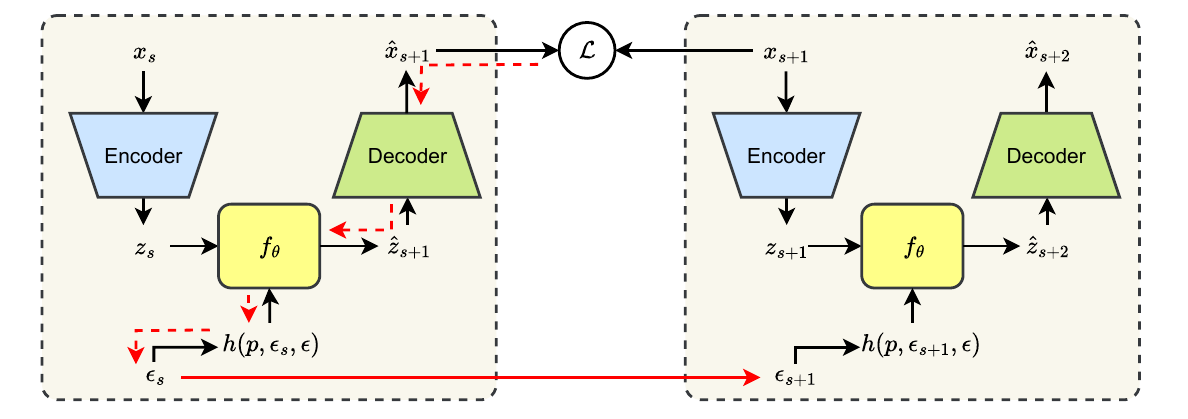}
   \caption{
   Overview of the proposed SAVi-DNO. Given noise $\epsilon_s$ and observation $x_s$ at time step $s$ of the video sequence, a future prediction $\hat{x}_{s+1}$ is generated through a diffusion model $f_\theta$, where the encoder and decoder map between pixel space $x$ and latent space $z$. After observing the next data $x_{s+1}$, we optimize the noise for the next prediction $\epsilon_{s+1}$, indicated by the red arrows. The function $h(p, \epsilon_s, \epsilon)$ adds some additional noise $\epsilon$ to $\epsilon_s$.
   }
   \label{fig:diff}
\end{figure*}

\section{Related Works}
\label{sec:relatedworks}

\myparagraph{Video Prediction.}
Early video prediction models were based on a deterministic paradigm \cite{shi2015convolutional, villegas2017decomposing, wang2017predrnn, wang2018predrnn++, byeon2018contextvp, dona2020pde, pourheydari2022taylor}. 
However, the uncertainty in the videos makes learning deterministic models difficult and causes them to collapse to make predictions corresponding to an average of possible futures. Stochastic models tackle this problem by modeling the inherent randomness present in the videos \cite{babaeizadeh2018stochastic, denton2018stochastic,akan2021slamp,castrejon2019improved,babaeizadeh2021fitvid,lee2018stochastic}.

Recently, following the success of diffusion models \cite{ho2020denoising, sohl2015deep, song2020score, song2019generative}, multiple video prediction method based on diffusion models have also been developed \cite{ho2020denoising,hoppe2022diffusion,voleti2022mcvd,lu2024vdt, zhang2024extdm}. MCVD \cite{voleti2022mcvd} proposes a masking strategy to train a single diffusion model capable of performing video prediction, generation, and interpolation tasks. In a similar approach, RaMViD \cite{hoppe2022diffusion} performs video prediction, infilling and upsampling. VDT \cite{lu2024vdt} integrates transformers with diffusion models for improved video prediction and generation. ExtDM \cite{zhang2024extdm} does video prediction by distribution extrapolation along temporal dimensions. PVDM \cite{yu2023video} proposes a triplane 2D autoencoder for more efficient video generation. SyncVP \cite{pallotta2025syncvp} builds on PVDM to do Multi-Modal video prediction. In a more recent work, Vista \cite{gao2025vista} extends Stable Video Diffusion \cite{blattmann2023stable} to a driving world model capable of action controllable generation in addition to video prediction.

\myparagraph{Diffusion Noise Optimization.}
Initial input noise to the sampling process of a diffusion model directly impacts the generated output. Given the fixed diffusion model parameters, one can search through the initial noise space to find an optimal noise based on a predefined criterion.
The work of \cite{qi2024not} hypothesizes that there is a fixed point noise based on diffusion sampling inversion and optimizes the noise to enforce this criterion for text-to-image generation. 
Similarly, to align text prompts better with the generated image, \cite{guo2024initno} uses a distribution optimization strategy and optimizes the distribution parameters instead of the noise itself. 
DOODL \cite{wallace2023end} uses the invertible reverse diffusion process from EDICT \cite{wallace2023edict} for memory-efficient optimization of the initial noise for the task of image editing. DNO \cite{karunratanakul2024optimizing} shows optimizing the noise with DDIM sampling is sufficient for various human motion generation tasks. Most recently, DITTO \cite{novack2024ditto} shows the effectiveness of the DNO with gradient checkpointing for music generation. We also follow a similar process to DNO in our work. 

\myparagraph{Learning from Continuous Videos Streams.}
Learning visual models from videos is usually done with the assumption of a finite prespecified set of videos, albeit a large one. These models are not made to benefit from the continuous observations from potentially infinite video streams. Among the works considering this setting, \cite{purushwalkam2022challenges} learns self-supervised image representations while going through a video stream. A minimum redundancy replay buffer is used during optimization to tackle the problem of overly correlated observed frames. In a similar approach, DoRA \cite{venkataramanan2024is} uses a multi-object tracking strategy to learn representations from a continuous video and use it for various downstream tasks. \cite{wang2023test} uses a masked autoencoding self-supervised training approach during test-time while processing a video stream to improve the main supervised task by just optimizing the self-supervised task. The aforementioned methods focus on learning image representations from the stream. In \cite{carreira2024learning} a simple video prediction is trained given a very long video and it is shown that given proper optimization strategies, learning from this continuous video performs on par with the standard learning methods. Similar to these methods, we aim to adapt to the video stream. However, we use a more advanced video prediction model while only adapting the initial noise instead of all the model parameters. Our goal is not to learn representation from the video stream but to perform better on the given sequence. 

%% file: sec/3_method.tex
\section{Video Prediction in Continuous Video Streams}
\myparagraph{Problem Statement.} Current approaches for video prediction are evaluated in an off-line setup. They are trained off-line on a dataset and then applied to video clips. Video prediction, however, has the advantage that new training data is freely available after each prediction. Furthermore, they are in practice applied to continuous video streams where the data distribution can change over time. We therefore investigate video prediction in continuous video streams, and we adapt a pre-trained diffusion model for video prediction continuously on the video stream. 

In practice, we work with short clips, where each clip $x_s \in \mathbb{R}^{H \times W \times S \times 3} $ consists of $S$ input frames. From this set of frames, we predict the next frames, \ie, $\hat{x}_{s+1} = f_\theta(x_s)$ using a pre-trained video prediction model $f_\theta$. As soon as we have observed the next clip $x_{s+1}$, we can compare it to the prediction $\hat{x}_{s+1}$ and update the parameters $\theta$. This makes it possible to gradually adapt $f_\theta$ to the characteristics of the current video sequence, potentially enhancing prediction performance as the model is continually updated with newly observed information. However, since $f_\theta$ is usually very large, fine-tuning $\theta$ is not very practical.

\myparagraph{Video Prediction Diffusion Model.} 
In the special case of Denoising Diffusion models, which are the common paradigm for video prediction, the prediction does not depend only on the observation $x_s$ and the parameters $\theta$, but also on the sampled noise $\epsilon_{s}$. Latent Denoising Diffusion models additionally employ an autoencoder to represent the data in a more compact and efficient way for the diffusion model, as illustrated in Figure~\ref{fig:diff}. During inference time, an Encoder $E$ transforms the input frames $x_{s}$ into latent representations $z_s = E(x_s)$. Then, the sampling process $f_\theta$ takes $z_s$ and sampled noise $\epsilon_{s}$ as input to predict the future latent representation  $\hat{z}_{s+1} = f_\theta(z_s, \epsilon_{s})$. Finally, a Decoder $D$ transforms the predicted future latent representation $\hat{z}_{s+1}$ to the pixel space $\hat{x}_{s+1} = D(\hat{z}_{s+1} )$. 

The function $f_\theta(z_s, \epsilon_{s})$ starts with sampling $\epsilon_s\sim\mathcal{N}(0,I)$ and setting $z_{s+1,T}=\epsilon_s$. $p_{\theta}(z_{s+1,t-1}|z_{s+1,t}, z_s)$ is then incrementally estimated for $t=T, \ldots, 1$. It is, however, sufficient to use the noise prediction function $\epsilon_{\theta}^t = \epsilon_{\theta}(z_{s+1,t},t,z_s)$ \cite{ho2020denoising}, which can be learned by optimizing $\| \epsilon - \epsilon_{\theta}(z_{s+1,t},t,z_s) \|^2_2$ for different values of $t$. Following DDIM~\cite{song2020denoising}, $z_{s+1,t-1}$ is generated by
\begin{equation}
\begin{split}
z_{s+1,t-1} &= \sqrt{\alpha_{t-1}} \left(\frac{z_{s+1,t} - \sqrt{1-\alpha_t}\epsilon_{\theta}^t}{\sqrt{\alpha_t}} \right) \\
&\quad + \sqrt{1-\alpha_t - \sigma_t^2(\eta)}\epsilon_{\theta}^t + \sigma_t(\eta) \epsilon,
\end{split}
\end{equation}
where $\alpha_t=\prod_{i=1}^t (1 - \beta_i)$, $\beta_t$ is pre-defined, and $\epsilon \sim\mathcal{N}(0,I)$. If $\sigma_t(\eta)$ is set to zero, the sampling except for the noise $\epsilon_s$ becomes deterministic. 

\myparagraph{Noise Optimization.} In order to improve the predictions of a pre-trained model $f_\theta$ on a continuous video stream, we do not optimize $\theta$ but $\epsilon_s$, every time we get a new observation $x_s$, \ie,
\begin{equation}\label{eq:opt}
 \epsilon^{*}_{s} = \arg\min_{\epsilon} \mathcal{L}\left(D\left(f_\theta(z_{s-1}, \epsilon\right)\right), x_{s}).
\end{equation}
This means that we optimize $\epsilon$ to minimize the loss between the prediction $\hat{x}_{s,\epsilon}=D(f_\theta(z_{s-1}, \epsilon))$ and observation $x_s$. 

As loss function, we calculate the difference between the prediction $\hat{x}_{s,\epsilon}$ and observation $x_{s}$
 \begin{equation}
\mathcal{L}_{pixel}(x_{s}, \hat{x}_{s,\epsilon}) =   \frac{1}{d_x}\| x_{s} - \hat{x}_{s,\epsilon} \|_1,
\end{equation}
where $d_x$ is the number of pixels, \ie, $d_x = H \times W \times S$. 

Additionally, we add a video feature loss $\mathcal{L}_{feature}$ to maintain semantic coherence of the predictions. We define the feature loss as 
\begin{equation}
\mathcal{L}_{feature}(x_{s}, \hat{x}_{s,\epsilon}; g_\phi) =  \frac{1}{d_f} \| g_\phi(x_{s}) - g_\phi(\hat{x}_{s,\epsilon}) \|_2^2
\end{equation}
where $g_\phi$ is a pre-trained video model with frozen parameters $\phi$ to extract high-level features across frames, and $d_f$ is the dimensionality of the features.

The final loss is then 
\begin{equation}\label{eq:loss}
\mathcal{L} = \mathcal{L}_{pixel}(x_{s}, \hat{x}_{s,\epsilon})  + \lambda \mathcal{L}_{feature}(x_{s}, \hat{x}_{s,\epsilon}; g_\phi),
\end{equation}
where $\lambda$ is a hyperparameter that controls the impact of the feature level loss. 

As an alternative, the difference between the prediction and observation can also be computed in the latent space, \ie,
\begin{equation}
\mathcal{L}_{latent}(z_{s}, \hat{z}_{s,\epsilon}) = \frac{1}{d_z} \| z_{s} - \hat{z}_{s,\epsilon} \|_1, 
\end{equation}
where $d_z$ is the dimension of the latent space. This, however, performs worse, as we show in our experiments. Nevertheless, it is an option for very large models with an expensive decoder like Vista \cite{gao2025vista}.   

While optimizing $\epsilon_s$ \eqref{eq:opt} improves the accuracy of the video prediction, it results in a deterministic prediction. In order to allow for some uncertainty in the prediction, we add additional noise to the optimized noise $\epsilon_s$, \ie,    
\begin{equation}
\begin{aligned}
h(p, \epsilon_s, \epsilon) &= \frac{p \epsilon_s + (1 - p) \epsilon}{\sqrt{p^2 + (1 - p)^2}}, 
\\
\hat{z}_{s+1} &= f_{\theta}(z_s, h(p, \epsilon_s, \epsilon)),
\end{aligned}
\end{equation}
where $\epsilon \sim \mathcal{N}(0,I)$ and $p\in [0,1]$ steers the randomness. The video prediction is deterministic if $p=1$, and the optimized noise is not used if $p=0$. The normalization is required to avoid an increase of the standard deviation. 
The steps of our approach are outlined in~\cref{alg:diffusion_noise_optimization_async}.  

\begin{algorithm}[t]
\caption{Sequence Adaptive Video Prediction with Diffusion Noise Optimization (SAVi-DNO)}
\label{alg:diffusion_noise_optimization_async}
\begin{algorithmic}[1]
\REQUIRE Pre-trained diffusion model $f_\theta$, encoder $E$, decoder $D$, parameter $p$, learning rate $l$, video stream $X$

\STATE $s \gets 1$
\STATE $\epsilon_s \sim \mathcal{N}(0,I)$
\WHILE{video is streaming}
    \STATE $x_{s} \gets X$
    \STATE $z_{s} \gets E(x_{s})$
    \STATE $\epsilon \sim \mathcal{N}(0,I)$
    \STATE $h(p, \epsilon_s, \epsilon) = \frac{p \epsilon_s + (1 - p) \epsilon}{\sqrt{p^2 + (1 - p)^2}}$
    \STATE $\hat{z}_{s+1,\epsilon_s} \gets f_{\theta}\left(z_s, h(p, \epsilon_s, \epsilon) \right)$
    \STATE $\hat{x}_{s+1,\epsilon_s} = D(\hat{z}_{s+1,\epsilon_s} )$
    \STATE $\nabla \gets \nabla_{\epsilon_s} \mathcal{L}(x_{s+1},\hat{x}_{s+1,\epsilon_s})$
    \STATE $\epsilon_{s+1} \gets \text{Optimizer}(\epsilon_s, \mathcal{L}, \nabla, l)$
    \STATE $s \gets s + 1$
\ENDWHILE
\end{algorithmic}
\end{algorithm}

%% file: sec/4_experiments.tex
\section{Experiments}

\subsection{Experiment Setup}

\myparagraph{Setting.}
In standard video prediction models, short clips from possibly long test videos are evaluated independently. Here, we consider a setting where test videos consist of long, continuous scenes rather than isolated clips.
Therefore, evaluation starts from the beginning of the long video and proceeds towards the end of it.

\myparagraph{Datasets.}
PVDM \cite{yu2023video} uses the SkyTimelapse \cite{xiong2018learning} and UCF-101 \cite{soomro2012ucf101} datasets for video generation evaluation. The corresponding pre-trained weights on both of these datasets are provided by the authors. We follow the same train and test splits as PVDM \cite{yu2023video} and other recent video generation methods \cite{skorokhodov2022stylegan,yu2022generating}.

Due to the availability of pre-trained models, we utilize the test sets for these two datasets. To ensure sufficient data for long-term evaluation, we filter out videos that cannot provide at least 10 consecutive observation-target future pairs.
In the end, we keep 1683 sequences out of 3783 for UCF-101 and 96 sequences out of 196 for the SkyTimelapse dataset.
The SkyTimelapse dataset has some relatively long sequences but does not involve very dynamic and complex scenarios. UCF-101, while being a more difficult dataset for video generation, has relatively short sequences. Therefore, we use Ego4D \cite{grauman2022ego4d} as additional dataset for our experiments. This dataset has very long sequences of challenging and dynamic scenes from an egocentric point of view. We take the minutes-long clips of Ego4D as the long sequences in our work. To minimize the similarity between test and train videos and simulate a more realistic scenario, we divide the videos based on the user identification of the recorded videos. We took a maximum of 3 videos from each user. Ultimately, we ended up with 1295 training videos and 267 evaluation videos, further split into 48 validation videos and 219 test videos. 

Additionally, we experiment with the Vista \cite{gao2025vista} foundation model on the validation set of the OpenDV-YouTube \cite{yang2024generalized} dataset given its pretrained weights. This dataset consists of long driving videos from diverse scenes. We use the first 20 minutes of the videos from this dataset.

For all datasets, we will provide scripts for optimization and evaluation. We will also release the source code upon acceptance.

\myparagraph{Evaluation.}
We use Structural Similarity Index Measure (SSIM) and Peak Signal-to-Noise Ratio (PSNR) as metrics for evaluating how accurate the video predictions are compared to the target frames. Additionally, Frechet Video Distance (FVD) \cite{unterthiner2018towards} is used to measure the quality of the predicted video frames. We calculate FVD based on all the predictions and the corresponding distribution of the target values.  

\myparagraph{Implementation Details.}
For all the datasets except OpenDV-YouTube, the frames are center-cropped and resized to $256\times256$. We use $320\times576$ resolution for OpenDV-YouTube. $S=16$ and $S=22$ are utilized for PVDM and Vista, respectively. PVDM predicts 16 frames given 16 observed input frames. A stride of 16 frames is used to move to the next location in the sequence. Vista predicts 22 frames given 3 frames and moves with a stride of 22. 
We use the pre-trained PVDM models for the SkyTimelapse and UCF-101 datasets, and the pretrained Vista model for OpenDV-Youtube. We trained PVDM on the Ego4D dataset using the public available source code.
The autoencoder was trained with a batch size of 8 for 288k iterations while the GAN loss was used in the last 30k iterations. The diffusion model was trained for 105k iterations with a batch size of 128 and a conditional probability of 0.7. 

For the noise optimization part, we use Adam \cite{kingma2013auto} with learning rates of 0.01, 0.01, 0.05, and 0.005 for Ego4D, UCF-101, SkyTimelapse, and OpenDV-YouTube, respectively. For PVDM, we do not use guidance during inference. Deterministic DDIM sampling ($\eta=0$) is used for the noise optimization experiments, while DDIM sampling with randomness ($\eta=1.0$) is used for sampling from the baseline PVDM model since PVDM works best in that setting.
We use DDIM with 10 sampling steps for the majority of the ablation experiments. Additionally, we report results for 50 sampling steps. Gradient checkpointing is used to reduce the memory consumption. 
While using Vista, we keep the default parameters for sampling except the sampling steps, which we set to 5.

We use a ResNet3D model \cite{tran2018closer} pretrained on Kinetics dataset \cite{kay2017kinetics} as the feature extraction network $g_{\phi}$.
$\lambda$ in \eqref{eq:loss} is set to (0.002, 0.012), (0.001, 0.0015), (0.0001, 0.0012) for (10 50) sampling steps on the Ego4D, SkyTimelapse and UCF-101 datasets, respectively.

\begin{table}[]
\centering
\begin{tabular}{lccc}
\toprule
Variant     &SSIM$\uparrow$&PSNR$\uparrow$& FVD$\downarrow$\\\midrule
PVDM w/o optimization          & 0.451 & 16.19 & 500.3  \\
$\mathcal{L}_{latent}$         & 0.478 & 16.81 & 517.6  \\
$\mathcal{L}_{latent} + noise$      & 0.467 & 16.49 & 486.8  \\
$\mathcal{L}_{pixel}$     & \textbf{0.491} & \textbf{17.10} & 535.1  \\
$\mathcal{L}_{pixel} + \mathcal{L}_{feature}$ & 0.485 & 17.08 & \textbf{463.9}  \\
$\mathcal{L}_{pixel} + \mathcal{L}_{feature} + noise$ & 0.485 & 17.02 & 466.3  \\
\hline
PVDM Inverse  & 0.391 & 15.05 & 190.1\\
\bottomrule
\end{tabular}
\caption{Analysis of the baselines and different variants of SAVi-DNO.}
  \label{tab:variants}
\end{table}

\begin{table}[]
\centering
\begin{tabular}{llccc}
\toprule
Variant     & $k$&SSIM$\uparrow$&PSNR$\uparrow$ & FVD$\downarrow$\\\midrule
Autoencoder  & & 0.745& 24.69& 47.7  \\\midrule
PVDM         & 1 & 0.451 & 16.19 & 500.3\\
PVDM (Best)  & 10 & 0.495 & 17.18 & 487.7\\
PVDM (Best)  & 20& 0.503& 17.37& 488.9  \\
\midrule
PVDM+SAVi-DNO & 1 & 0.485 & 17.02& \textbf{466.3 }\\
\bottomrule
\end{tabular}
\caption{Upper bounds considering the PVDM method and its autoencoder. The first row reports the reconstruction error of the autoencoder without video prediction. Rows 2-4 report the best results out of $k$ samples, where the best sample is selected based on the ground-truth.}
  \label{tab:upperbound}
\end{table}

\begin{figure*}[t]
    \centering
   \includegraphics[trim=0 0 0 0, width=0.9\linewidth]{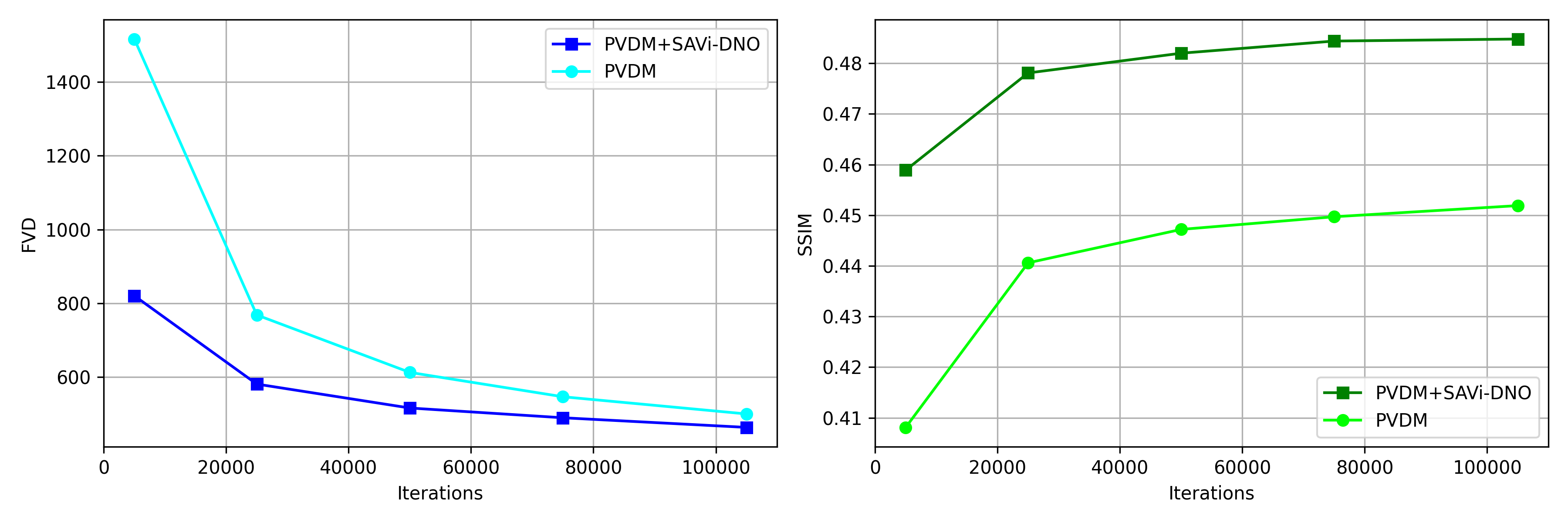}
   \caption{Performance comparison of SAVi-DNO and base PVDM with varying amounts of pretraining for the PVDM diffusion model.}
   \label{fig:iter}
\end{figure*}
\subsection{Ablations}
In this part, we perform various calculations on the validation part of the Ego4D dataset to analyze the reliability of the noise optimization method for improving the video prediction performance.

The results for the main components of our approach in addition to the baselines are provided in~\cref{tab:variants}. In the following, we describe different baselines and variants of our method.

\textbf{PVDM w/o optimization.} Base PVDM model without any noise optimization.

\textbf{PVDM Inverse.} Since we are looking for a suitable input noise for the current sequence, we can run DDIM Inversion given the observed target value $z_{s+1}$ to obtain the approximate noise $\epsilon^{inverse}_s$. We can use $\epsilon^{inverse}_s$ at step $s+1$ to make the next prediction. We use DDIM Inversion as a baseline in addition to just running the base PVDM model. 

$\boldsymbol{\mathcal{L}_{latent}.}$ Here, we only use the $\mathcal{L}_{latent}$ to optimize the input noise. This variant is particularly beneficial for large models due to their increased memory requirements. 

$\boldsymbol{\mathcal{L}_{latent} + noise.}$ The random noise interpolation function $h(p, \epsilon_s, \epsilon)$ is used to introduce controllable randomness while only optimizing the $\mathcal{L}_{latent}$. 

$\boldsymbol{\mathcal{L}_{pixel}.}$ This variant corresponds to the case where we use the decoder $D$ to first decode the latents $\hat{z}_{s+1}$ to the frames $\hat{x}_{s+1}$ then calculate the loss $\mathcal{L}_{pixel}$ on pixels.

$\boldsymbol{\mathcal{L}_{pixel} + \mathcal{L}_{feature}.}$ The same scenario as the previous variant with the added feature loss $\mathcal{L}_{feature}$.

$\boldsymbol{\mathcal{L}_{pixel} + \mathcal{L}_{feature}+ noise.}$ 
Final version of the model which is optimized with the $\mathcal{L}_{pixel} + \mathcal{L}_{feature}$ and includes the random noise interpolation.

 \myparagraph{Variant Analysis.} Here, we explain the results from ~\cref{tab:variants} and analyze the baselines and each component of our model. Interestingly, the PVDM Inverse is significantly worse than PVDM. One explanation for this outcome is that the DDIM Inverse noise distribution is different from the actual noise distribution possibly having correlated elements in the inverse noise which may not be suitable for the next prediction. Additionally, the trajectory of DDIM Inverse noise in the video sequence may not be sufficiently smooth for it to be used for the next prediction. 
We observe a far better FVD value of 190.1 for PVDM Inverse due to similar predictions to the last timestep, which is not useful by itself without an improved SSIM value.
Next, we see that $\mathcal{L}_{latent}$ improves SSIM and PSNR noticeably (+0.027 and +0.62). However, a degradation in the FVD metric is observed. Random noise interpolation with $p=0.5$ fixes the FVD degradation at the cost of reduced relative improvement of SSIM and PSNR but still a better performance than base PVDM is achieved. 
$\mathcal{L}_{pixel}$ offers even more improvements on SSIM and PSNR compared to PVDM (+0.040 and +0.91) but the FVD becomes even higher than $\mathcal{L}_{latent}$ which is not desirable. Finally, we can see that the feature loss in the $\mathcal{L}_{pixel} + \mathcal{L}_{feature}$ variant gives the best FVD, while still offering significant improvements based on SSIM and PSNR (+0.034, +0.89) compared to PVDM. This variant performs best in terms of metrics, but it does not take into account any randomness. Adding random noise interpolation with $p=0.9$ to the final variant does not reduce the performance, while also considering some level of uncertainty. 

\begin{table}[b]
\centering
\begin{tabular}{lccc}
\toprule
Variant     &SSIM$\uparrow$&FVD$\downarrow$& Time (s)\\\midrule
PVDM                     & 0.451 & 500.3 & 1.37  \\ 
\hline
PVDM + diff opt (1)      & 0.457 & 501.1 & 1.74  \\
PVDM + diff opt (10)     & 0.457 & 497.3 & 2.11  \\
PVDM + diff opt (20)     & 0.458 & 496.3 & 2.85  \\
PVDM + diff opt (100)    & 0.460 & 483.0 & 8.79  \\
\hline
PVDM + ours              & 0.485 & 466.3 & 3.57  \\
\hline
PVDM + ours every 2      & 0.480 & 449.4 & 2.47  \\
PVDM + ours every 5      & 0.473 & 440.5 & 1.81  \\
PVDM + ours every 10     & 0.468 & 435.4 & 1.60  \\
\bottomrule
\end{tabular}
\caption{The effect of full model optimization with diffusion loss compared to noise optimization considering the computation time calculated on an Nvidia 3090 GPU. Diff opt ($n$) shows full model optimization repeated $n$ times at each sequence step.
}
  \label{tab:alternative}
\end{table}

\myparagraph{Performance Upperbound.}
In~\cref{tab:upperbound}, we compare the performance of our approach with an oracle that selects among multiple predictions of PVDM for each instance the one with best SSIM and PSNR. This gives an indication how close our approach is to the best prediction that PVDM can generate. We observe that running PVDM 10 and 20 times leads to SSIM values of 0.495 and 0.503, respectively, which is significantly higher than running PVDM once 0.451. Interestingly, our approach with an SSIM value of 0.485 can recover a good portion of the performance of the oracle. This is a strong indication of the success of our approach to find an optimal noise value for the current input given the sequence of previous observations. We also include the ground truth reconstruction performance by the autoencoder to show the upper bound given this autoencoder. Note that the autoencoder encodes and decodes the ground-truth, and a prefect prediction model could not be better than the reconstruction of its autoencoder. 

\myparagraph{Diffusion Loss Optimization.}
A more costly alternative to diffusion noise optimization is training the model weights with the diffusion noise prediction loss while observing the continuous sequence. In ~\cref{tab:alternative}, we compare the performance and computation time of adapting the model weights to our approach of optimizing the noise using the PVDM model and Ego4D dataset. Optimizing the model parameters with diffusion loss adapts to the sequence very slowly. It can be observed that optimizing the model parameters 100 times per observation yields worse SSIM and FVD (-0.008, +47.6) than our method, even if the noise optimization is performed once every 10 observations, while being significantly slower (8.79 vs 1.60 seconds).  
Considering the computational efficiency, PVDM requires 1.37 seconds. Our method adds 2.2s, where 2.18s are for backpropagation and 0.02s for the additional feature loss. If we perform the noise optimization only every $k$ steps, we achieve a trade-off between computation time and accuracy. For $k=10$, the computational overhead is very small, but SSIM and FVD are still improved.


\begin{figure}[t]
\centering
   \includegraphics[trim=0 0 0 0, width=1.0\linewidth]{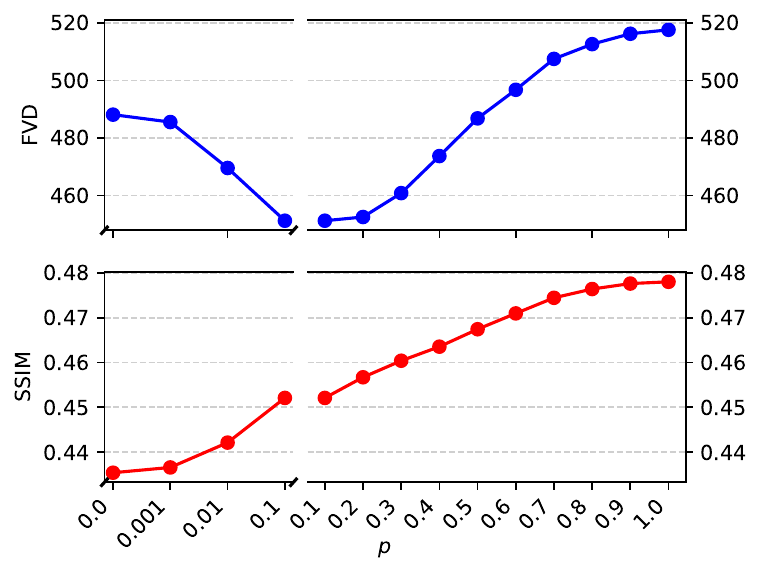}
   \caption{The impact of noise interpolation parameter $p$ on FVD and SSIM of the $\mathcal{L}_{latent} + noise$ variant of the method on the validation set of Ego4D.}
   \label{fig:p}
\end{figure}

\begin{table*}[]
\centering
\begin{tabular}{llccc|ccc|ccc}
\toprule
\multirow{2}{*}{Steps} &  \multirow{2}{*}{Method}    & \multicolumn{3}{c}{Ego4D}                 & \multicolumn{3}{c}{UCF101}               & \multicolumn{3}{c}{SkyTimelapse}\\\cmidrule(lr){3-5}\cmidrule(lr){6-8} \cmidrule(lr){9-11}
                    &              & FVD$\downarrow$&SSIM$\uparrow$&PSNR$\uparrow$ & FVD$\downarrow$&SSIM$\uparrow$&PSNR$\uparrow$ & FVD$\downarrow$&SSIM$\uparrow$&PSNR$\uparrow$ \\\midrule
\multirow{2}{*}{10} & PVDM         & 427.1& 0.463 & 16.26 & 763.3& 0.638 & 18.93 & 179.8 & 0.785 & 22.72 \\
                    & PVDM+SAVi-DNO    & \textbf{384.5}& \textbf{0.493} & \textbf{17.01} & \textbf{753.1}& \textbf{0.668} & \textbf{19.90} & \textbf{163.1} & \textbf{0.829} & \textbf{24.95} \\\midrule
\multirow{2}{*}{50} & PVDM         & 244.6& 0.436 & 15.71 & 545.1& 0.609 & 18.18 & 119.8& 0.764 & 21.87 \\
                    & PVDM+SAVi-DNO    & \textbf{231.0}& \textbf{0.464}& \textbf{16.44}& \textbf{543.4}& \textbf{0.650}& \textbf{19.42}& \textbf{115.1}& \textbf{0.822}& \textbf{24.77}\\ 
\bottomrule
\end{tabular}
\caption{Results on the test sets of Ego4D, UCF101, and SkyTimelapse.}
  \label{tab:main}
\end{table*}

\myparagraph{Training Budget.} In~\cref{fig:iter}, we show the effect of SAVi-DNO based on the amount of training iterations. Interestingly, SAVi-DNO makes the performance of the model in different iterations more robust and reduces the performance difference even with the models that were trained for as few as 5k iterations. Using SAVi-DNO with the PVDM trained for 50k iterations performs on par with PVDM trained for 105k iterations in terms of FVD and outperforms it significantly based on the SSIM metric. Therefore, training cost reduction can also be considered as a possible benefit of SAVi-DNO. 

\begin{figure}[t]
\centering
   \includegraphics[trim=0 0 0 0, width=1.0\linewidth]{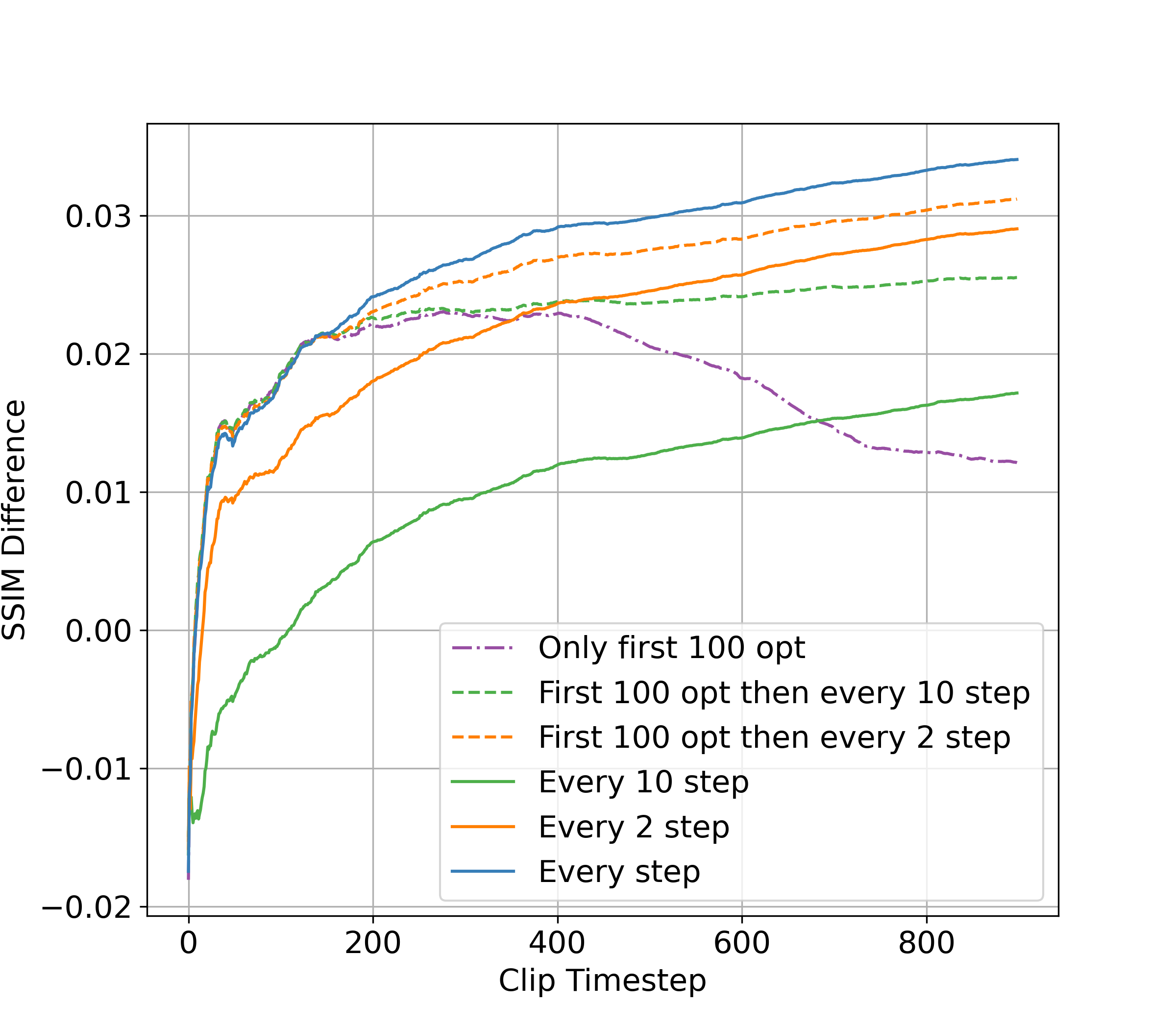}
   \caption{SSIM evaluation progression over video length on the Ego4D validation split. Every point on the graph indicates the difference in the SSIM metric of the SAVi-DNO variant with PVDM. 
   Every value on the x-axis means the model was evaluated until the first x clips.
   }
   \label{fig:ssim}
\end{figure}

\myparagraph{Performance Over the Sequence.} \cref{fig:ssim} shows the performance gain of the model compared to PVDM along the length of the sequence. We can see that for the default SAVi-DNO setting (Every step) the positive SSIM gap relative to PVDM consistently keeps improving the longer the videos are. Less frequent optimization steps also show a similar pattern of consistent improvement but with a meaningful gap compared to optimizing every step. Additionally, we can see that an initial warmup phase of optimizing first 100 times every step can provide a head start to less frequent optimization variants to reduce the gap to a setup where we optimize every step. However, if we do not optimize after the warmup phase, the performance gain decreases. It should be noted that the initial negative performance gain is due to the reason that we use deterministic DDIM compared to the better performing DDIM with $\eta=1$ in PVDM. 

\myparagraph{Random noise interpolation.} The value of $p$ controls the amount of random noise interpolation. The effect of this value is demonstrated in~\cref{fig:p}. We conduct this ablation with the $\mathcal{L}_{latent} + noise$ version of the method to rule out the effect of other factors like the $\mathcal{L}_{feature}$. The closer the $p$ is to $1$, we get higher SSIM, but the FVD increases as well. Lower value of $p$ offers lower SSIM increase while improving FVD before increasing it again in the regions closer to $0$. At $0$, random noise completely replaces the optimized noise and the FVD and SSIM values equal PVDM with $\eta=0$. The explanation for this behaviour is that some level of noise optimization actually improves both SSIM and FVD. However, more focus on the optimized noise can harm the diversity of predictions. It improves SSIM at the cost of higher FVD.

\begin{table}[]
\centering
\begin{tabular}{lccc}
\toprule
Method      &SSIM$\uparrow$&PSNR$\uparrow$ & FVD$\downarrow$\\\midrule
Vista& 0.498& 15.87&  974.2 \\
Vista+SAVi-DNO& 0.509 & 16.17 & \textbf{945.5 }\\
\bottomrule
\end{tabular}
\caption{Results for the VISTA method on the validation set of the OpenDV-Youtube dataset.}
  \label{tab:main_vista}
\end{table}

\begin{figure*}[t]
\centering
   \includegraphics[trim=0 0 0 0, width=1.0\linewidth]{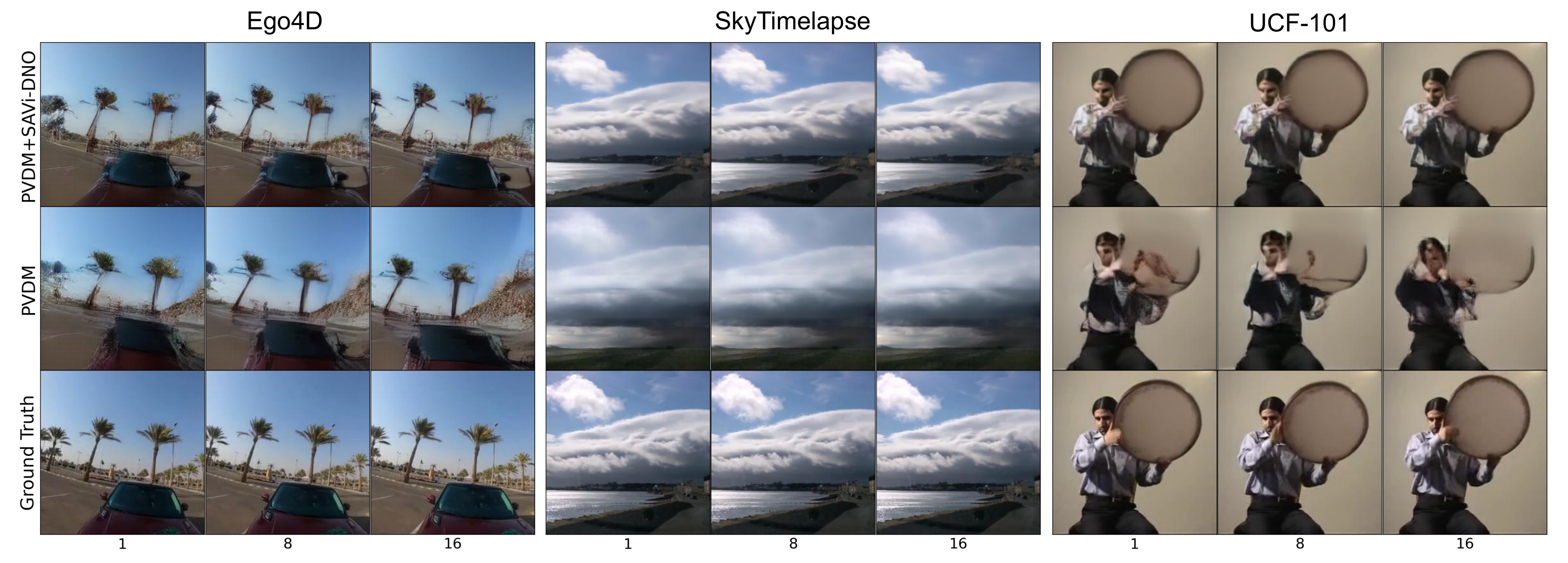}
   \caption{Qualitative results on Ego4D, UCF101 and SkyTimelapse datasets using PVDM and PVDM+SAVi-DNO.}
   \label{fig:qual}
\end{figure*}
\begin{figure*}[t]
\centering
   \includegraphics[trim=20 10 15 0, width=0.95\linewidth]{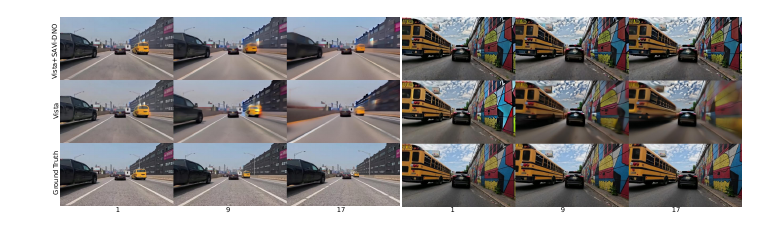}
   \caption{Qualitative results on OpenDV-Youtube dataset using Vista and Vista+SAVi-DNO.}
   \label{fig:qual_vista}
\end{figure*}

\subsection{Results}

\subsubsection{Quantitative Results}
The results for SAVi-DNO compared to PVDM on the test splits of Ego4D, UCF-101, and SkyTimelapse are provided in~\cref{tab:main}. On all datasets, we see consistent improvements based on all the metrics across various DDIM sampling steps. Generally, the challenging nature of the Ego4D dataset is evident from the SSIM and PSNR metrics. SAVi-DNO can enhance the performance of PVDM for all datasets and metrics. 

Additionally, we report the results of the larger Vista model on the OpenDV-Youtube dataset in~\cref{tab:main_vista}. Due to the heavier memory consumption of this model, we use the $L_{latent} + noise$ variant showing the benefit of this variant for larger models. We observe higher SSIM and PSNR when using SAVi-DNO for Vista, while simultaneously decreasing FVD. 

\subsubsection{Qualitative Results}
\cref{fig:qual} illustrates the video prediction results for the Ego4D, SkyTimelapse and UCF-101 datasets. Generally, we can see a clear improvement by SAVi-DNO in the fidelity of the predicted frames across all three datasets. On the Ego4D sample, the camera view change caused by the egomotion is better captured compared to the large turn to the right predicted by PVDM.  
In addition to more accurate outputs for the clouds and the ground in the SkyTimelapse predictions, the movement of the smaller cloud on the top left of the SkyTimelapse frames is more consistent with ground truth for SAVi-DNO compared to PVDM. 
PVDM fails to consistently predict the shape of the musical instrument and the details of the person in the sample from the UCF-101 dataset, which is not the case when using SAVi-DNO. \cref{fig:qual_vista} shows similar improvements on OpenDV-Youtube dataset when SAVi-DNO is applied to the Vista model. The presence of the black car truck is better predicted with SAVi-DNO with sharper predictions in both cases. More qualitative samples are provided in the supplementary material.

%% file: sec/5_conclusion.tex
\section{Conclusion}
In this work, we introduced Sequence Adaptive Video Prediction with Diffusion Noise Optimization (SAVi-DNO), a video prediction model adaptation method for continuous video streams. We introduced a new setting for adapting and evaluating video prediction models in long continuous videos. We evaluated SAVi-DNO on four datasets and demonstrated that it improves the video prediction of the corresponding diffusion models PVDM and Vista. 

%% file: sec/X_suppl.tex
\clearpage
\setcounter{page}{1}
\maketitlesupplementary

\section{Additional Experiments}
\label{sec:exp2}

\myparagraph{Sampling Steps.} The effect of SAVi-DNO on different DDIM sampling steps is shown in ~\cref{tab:steps}. There is a consistent improvement in all three sampling step numbers of 10, 20, and 50 based on all the evaluation metrics compared to PVDM. However, with higher sampling steps, we see a drop in the performance of pairwise calculated metrics (SSIM, PSNR) in the base PVDM model. The significantly lower FVD of 50 steps to 20 and that of 10 steps means that the predictions have better quality and diversity, but may not necessarily match the target values in pixel-to-pixel comparison. Nevertheless, we consistently have relative improvements with SAVi-DNO to the base PVDM. 
\begin{table}[]
\centering
\begin{tabular}{llccc}
\toprule
Steps               & Method     &SSIM$\uparrow$&PSNR$\uparrow$& FVD$\downarrow$\\\midrule
\multirow{2}{*}{10} & PVDM          & 0.451 & 16.19 & 500.3\\
                    \cmidrule{2-5}
                    & + SAVi-DNO  & \textbf{0.485} & \textbf{17.02}& \textbf{466.3 }\\\midrule
\multirow{2}{*}{20} & PVDM          & 0.440 & 15.92 & 366.7\\
                    \cmidrule{2-5}
                    & + SAVi-DNO  & \textbf{0.471}& \textbf{16.74}& \textbf{347.6}\\ \midrule
\multirow{2}{*}{50} & PVDM          & 0.426 & 15.66 & 287.0\\
                    \cmidrule{2-5}
                    & + SAVi-DNO  & \textbf{0.458}& \textbf{16.44}& \textbf{280.3}\\ 
\bottomrule
\end{tabular}
\caption{The effect of the DDIM sampling steps.}
  \label{tab:steps}
\end{table}

\begin{figure}[t]
\centering
   \includegraphics[trim=0 0 0 0, width=1.0\linewidth]{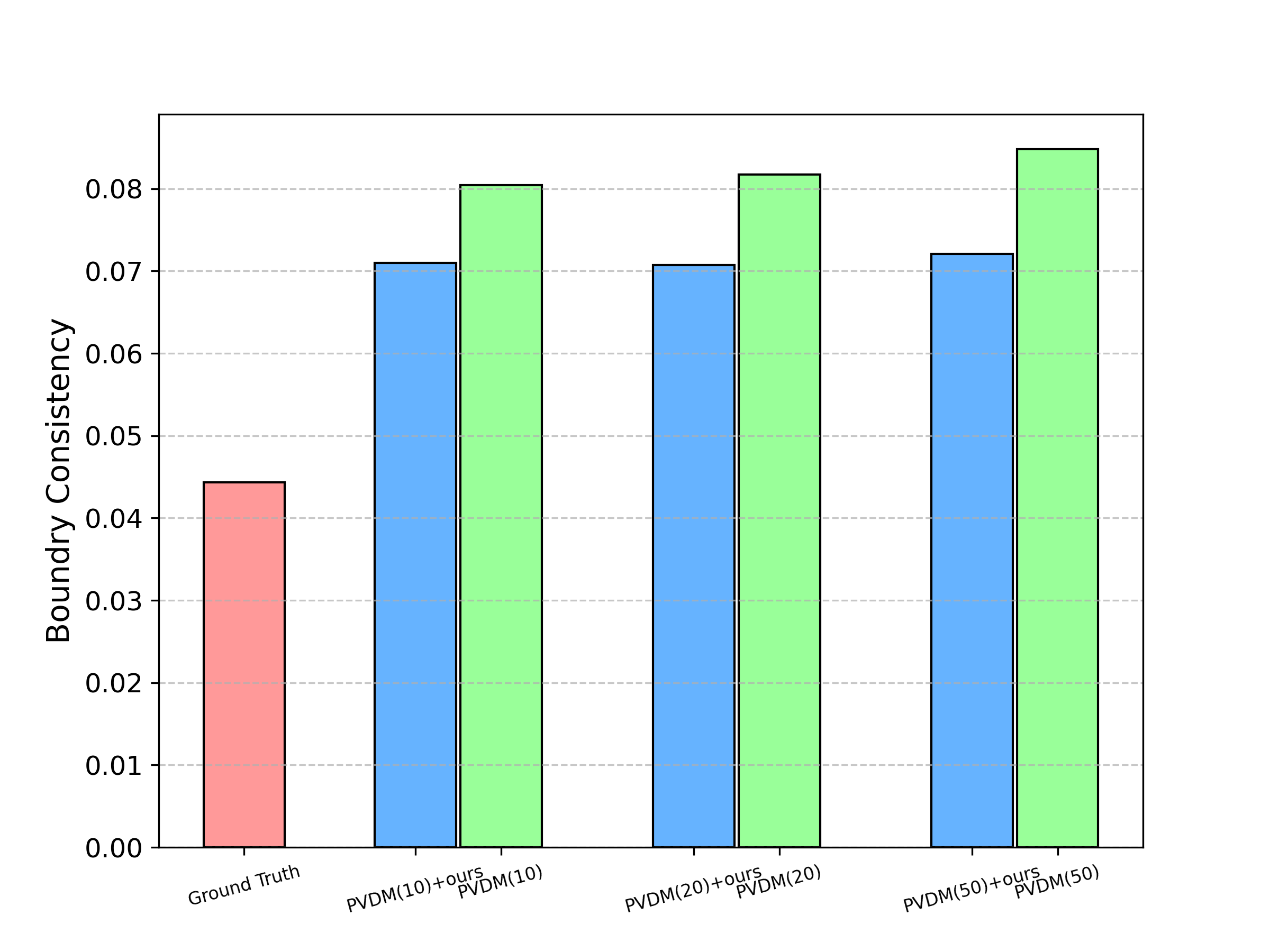}
   \caption{Comparison of the boundary consistency metric.}
   \label{fig:consistency}
\end{figure}

\myparagraph{Boundary Consistency.} We calculate the average pixelwise absolute error between the last condition frame and first prediction frame as a measure of consistent and smooth transition between the condition clip and the prediction. In ~\cref{fig:consistency}, we can see that in all the sampling steps on Ego4D validation set, the error is lower while using the SAVi-DNO and closer to the original difference of the two boundry frames. This means the transition is smoother and more consistent.

\myparagraph{Impact of $\eta$.} In ~\cref{tab:sigma}, we compare the impact of $\eta$ in DDIM sampling in PVDM and SAVi-DNO. In the original PVDM implementation $\eta=1$ is used which also performs better here. Therefore, we used $\eta=1$ for the experiments with PVDM. Using $\eta=0$ would result in an even worse performance compared to SAVi-DNO. When using SAVi-DNO with $\eta=1$, the additional random noise during the denoising steps of the sampling process alters the deterministic flow of operations required for calculating gradients. Consequently, the performance is degraded for pair-wise metrics. Therefore, we opt for the fully deterministic DDIM with $\eta=0$ while using SAVi-DNO. 
\begin{table}[]
\centering
\begin{tabular}{lcccc}
\toprule
Variant &$\eta$ &SSIM$\uparrow$&PSNR$\uparrow$ & FVD$\downarrow$\\\midrule
PVDM          &0 & 0.435 & 15.80 & 488.12 \\
PVDM+SAVi-DNO &0 & \textbf{0.485} & \textbf{17.02}& 466.32\\\midrule
PVDM          &1 & 0.451 & 16.19 & 500.30  \\
PVDM+SAVi-DNO &1 & 0.464 & 16.44 & \textbf{452.92}  \\
\bottomrule
\end{tabular}
\caption{Impact of DDIM $\eta$ on the Ego4D validation results while using 10 sampling steps.}
  \label{tab:sigma}
\end{table}

\myparagraph{Impact of $\lambda$.} The impact of hyperparameter $\lambda$ of the feature loss is analyzed in ~\cref{tab:lambda} for the $\mathcal{L}_{pixel} + \mathcal{L}_{feature}$ variant of the model to rule out the impact of random noise interpolation. Edge cases of not using the feature loss and only using the feature loss fail to provide a balance between FVD and pixel based metrics of SSIM and PSNR. We use a small value of $\lambda$ to have a trade-off between the metrics.

\begin{table}[]
\centering
\begin{tabular}{lcccc}
\toprule
Variant &$\lambda$ &SSIM$\uparrow$&PSNR$\uparrow$ & FVD$\downarrow$\\\midrule
$\mathcal{L}_{pixel}$                                 & -     & \textbf{0.491} & \textbf{17.10} & 535.1  \\
$\mathcal{L}_{pixel} + \lambda \mathcal{L}_{feature}$ & 0.002 & 0.485 & 17.08 & 463.9 \\
$\mathcal{L}_{pixel} + \lambda \mathcal{L}_{feature}$ & 0.005 & 0.476 & 16.91 & 414.2 \\
$\mathcal{L}_{pixel} + \lambda \mathcal{L}_{feature}$ & 0.01  & 0.468 & 16.76 & 370.0 \\
$\mathcal{L}_{pixel} + \lambda \mathcal{L}_{feature}$ & 0.1   & 0.440 & 16.10 & 296.5 \\
$\mathcal{L}_{feature}$                               & -     & 0.421 & 15.32 & \textbf{274.0} \\
\bottomrule
\end{tabular}
\caption{Impact of $\mathcal{L}_{feature}$ hyperparameter $\lambda$ on the Ego4D validation set with 10 DDIM sampling steps.}
  \label{tab:lambda}
\end{table}

\myparagraph{Dataset Transfer.} In order to see the applicability of SAVi-DNO given a pretrained network on an unrelated data distribution to the test data, we perform experiments given all the possible pairs of PVDM pretrained model and test data in ~\cref{tab:pretrain}. We can see comparable performance in most cases, except when the SkyTimelapse dataset is the training dataset. This is expected since this dataset is very limited to videos from sky and does not include enough variation to learn a general enough model to be adaptable to other datasets. On the contrary, we can see that the model trained on the highly complicated data of Ego4D can outperform the model trained on the SkyTimelapse model while using SkyTimelapse as test data. 
Generally, we can see that the noise optimization with SAVi-DNO can achieve comparable performance to the base method without training on the data from the same distribution as the test data.

\begin{figure}[t]
\centering
   \includegraphics[trim=0 0 0 0, width=1.0\linewidth]{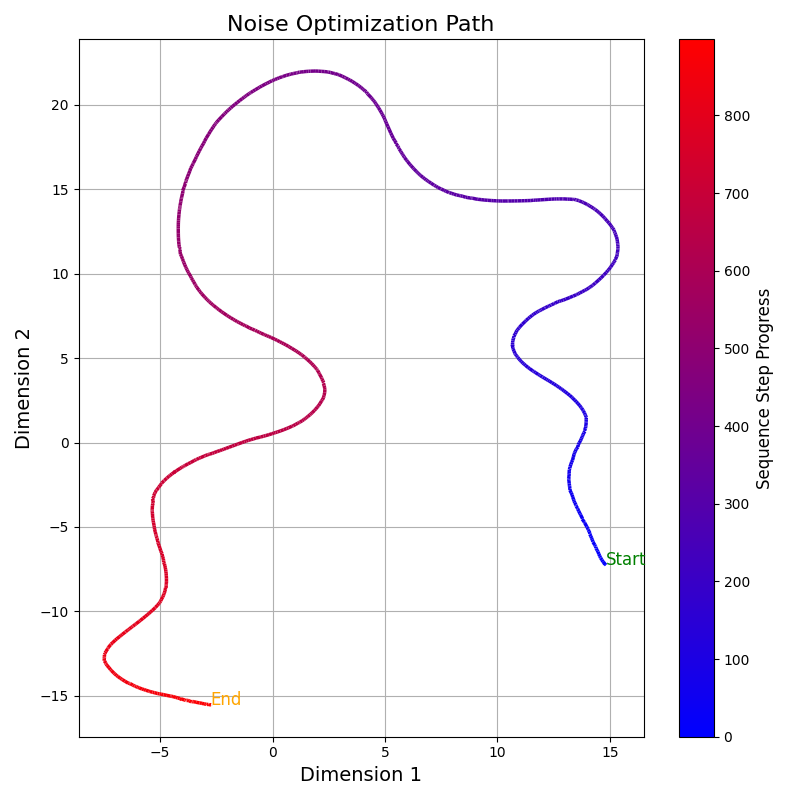}
   \caption{Noise optimization path visualized in two dimensions with UMAP.}
   \label{fig:optpath}
\end{figure}

\begin{table*}[]
\centering
\begin{tabular}{llccc|ccc|ccc}
\toprule
\multirow{2}{*}{Train Data} &  \multirow{2}{*}{Method}    & \multicolumn{3}{c}{Ego4D}                 & \multicolumn{3}{c}{UCF101}               & \multicolumn{3}{c}{SkyTimelapse}\\\cmidrule(lr){3-5}\cmidrule(lr){6-8} \cmidrule(lr){9-11}
                    &              & SSIM$\uparrow$&PSNR$\uparrow$ &FVD$\downarrow$& SSIM$\uparrow$&PSNR$\uparrow$ &FVD$\downarrow$& SSIM$\uparrow$&PSNR$\uparrow$ &FVD$\downarrow$\\\midrule
\multirow{2}{*}{Ego4D} & PVDM         & 0.451 & 16.19 & 500.3  & 0.627& 19.17& 1089.9& 0.785& 23.29& 151.1\\
                    &  +SAVi-DNO    & \textbf{0.485}& \textbf{17.02}& \textbf{466.3}& 0.643& 19.64& 1077.9& 0.795& 24.53& \textbf{147.0}\\\midrule
\multirow{2}{*}{UCF101} & PVDM         & 0.432& 15.61& 718.3& 0.638 & 18.93 & 763.3& 0.771& 22.34& 193.9\\
                    &  +SAVi-DNO    & 0.484& 16.91& 560.0& \textbf{0.668}& \textbf{19.90}& \textbf{753.1}& 0.820& 24.86& 188.4\\ \midrule
\multirow{2}{*}{SkyTimelapse} & PVDM         & 0.411& 14.26& 1716.7& 0.541& 16.47& 2446.5& 0.785 & 22.72 & 179.8 \\
                    &  +SAVi-DNO    & 0.460& 16.09& 1121.7& 0.602& 18.27& 2086.4& \textbf{0.829}& \textbf{24.95}& 163.1\\ 
\bottomrule
\end{tabular}
\caption{Results on Ego4D (validation), UCF101, and SkyTimelapse while using models trained on a different dataset than the test data.}
  \label{tab:pretrain}
\end{table*}

\myparagraph{Optimization Path.} In Figure~\ref{fig:optpath}, the noise optimization path is visualized using UMAP \cite{mcinnes2018umap} in two dimensions. Noise smoothly moves to different regions of the noise space in order to adapt to the new observations from the sequence. It can be considered that the optimal noise is found locally and changes based on different parts of the sequence.

\section{Qualitative Results}
\label{sec:qual2}

\begin{figure*}[t]
\centering
   \includegraphics[trim=0 0 0 0, width=1.0\linewidth]{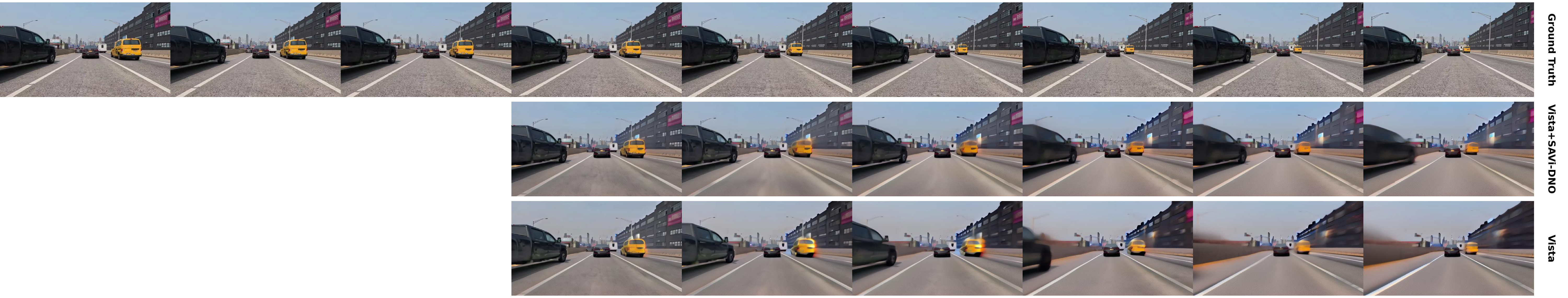}
   \caption{OpenDV-Youtube qualitative sample.}
   \label{fig:odv1}
\end{figure*}

\begin{figure*}[t]
\centering
   \includegraphics[trim=0 0 0 0, width=1.0\linewidth]{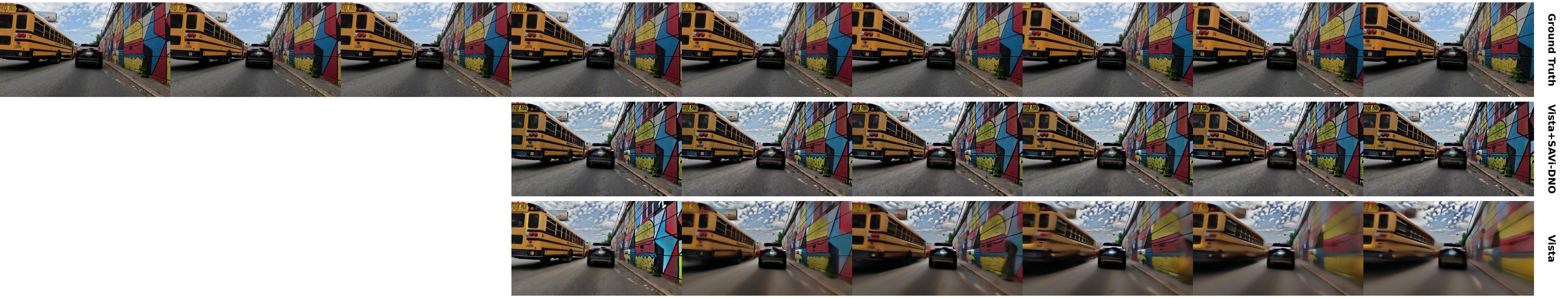}
   \caption{OpenDV-Youtube qualitative sample.}
   \label{fig:odv2}
\end{figure*}

\begin{figure*}[t]
\centering
   \includegraphics[trim=0 0 0 0, width=1.0\linewidth]{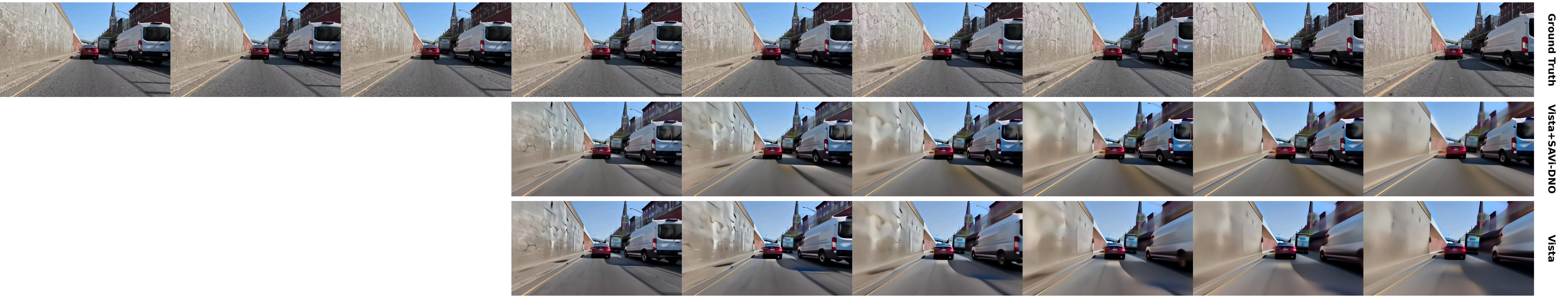}
   \caption{OpenDV-Youtube qualitative sample.}
   \label{fig:odv3}
\end{figure*}

\begin{figure*}[t]
\centering
   \includegraphics[trim=0 0 0 0, width=1.0\linewidth]{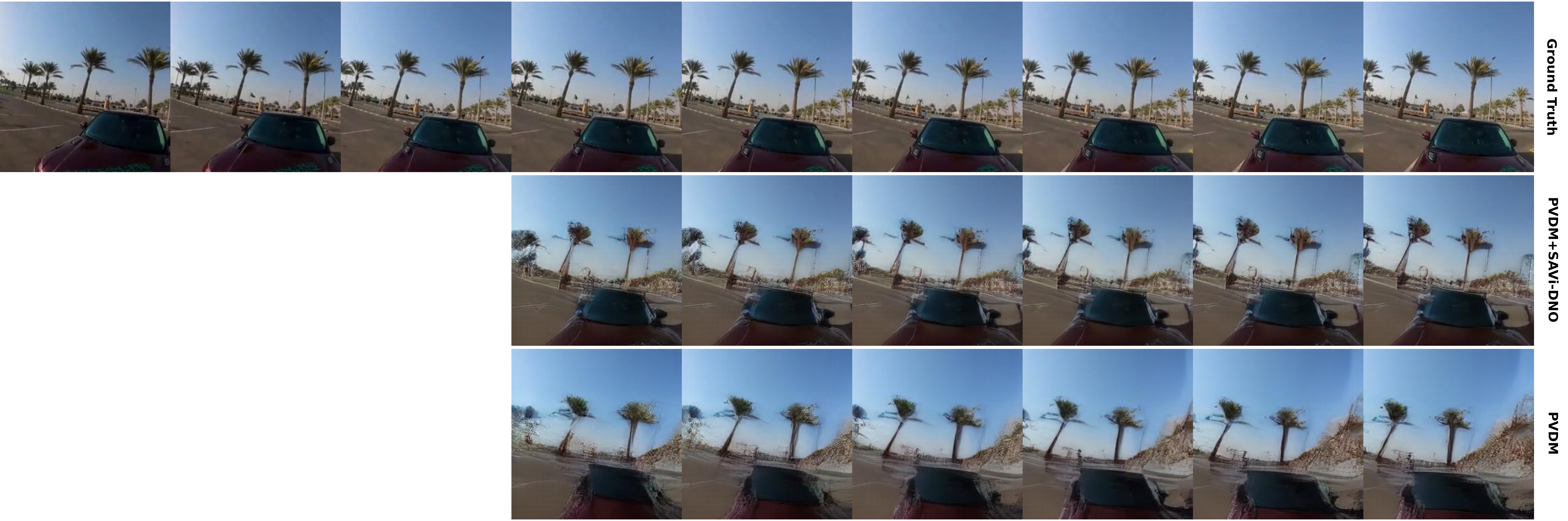}
   \caption{Ego4D qualitative sample.}
   \label{fig:ego4d1}
\end{figure*}

\begin{figure*}[t]
\centering
   \includegraphics[trim=0 0 0 0, width=1.0\linewidth]{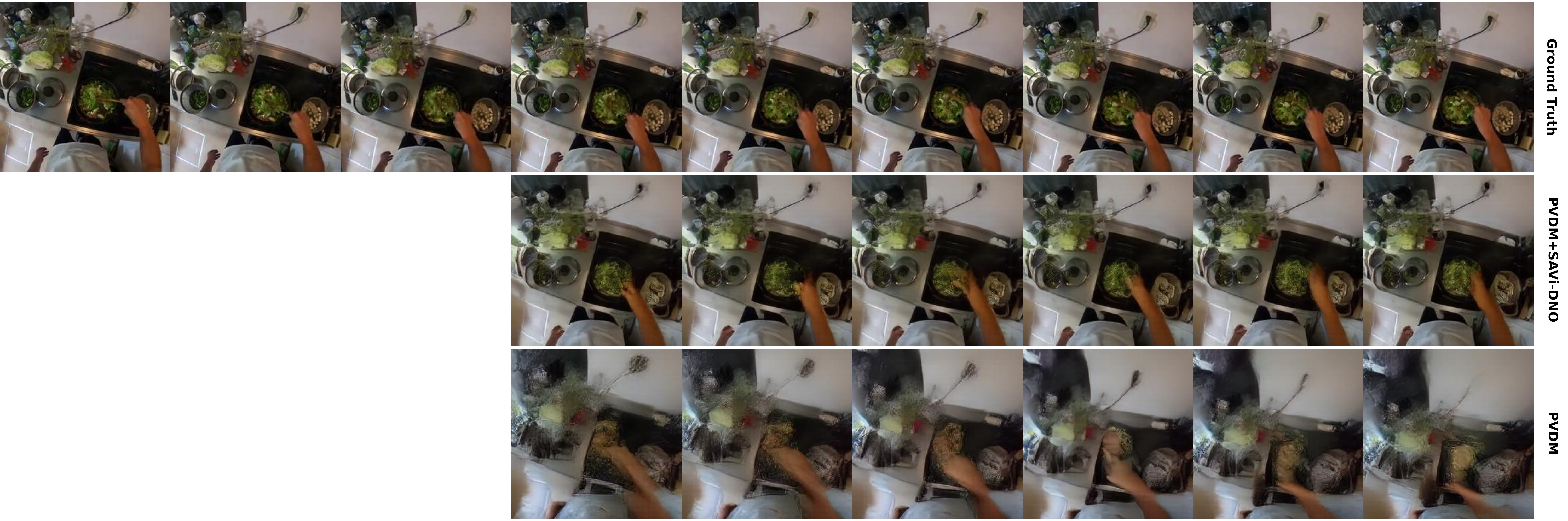}
   \caption{Ego4D qualitative sample.}
   \label{fig:ego4d2}
\end{figure*}

\begin{figure*}[t]
\centering
   \includegraphics[trim=0 0 0 0, width=1.0\linewidth]{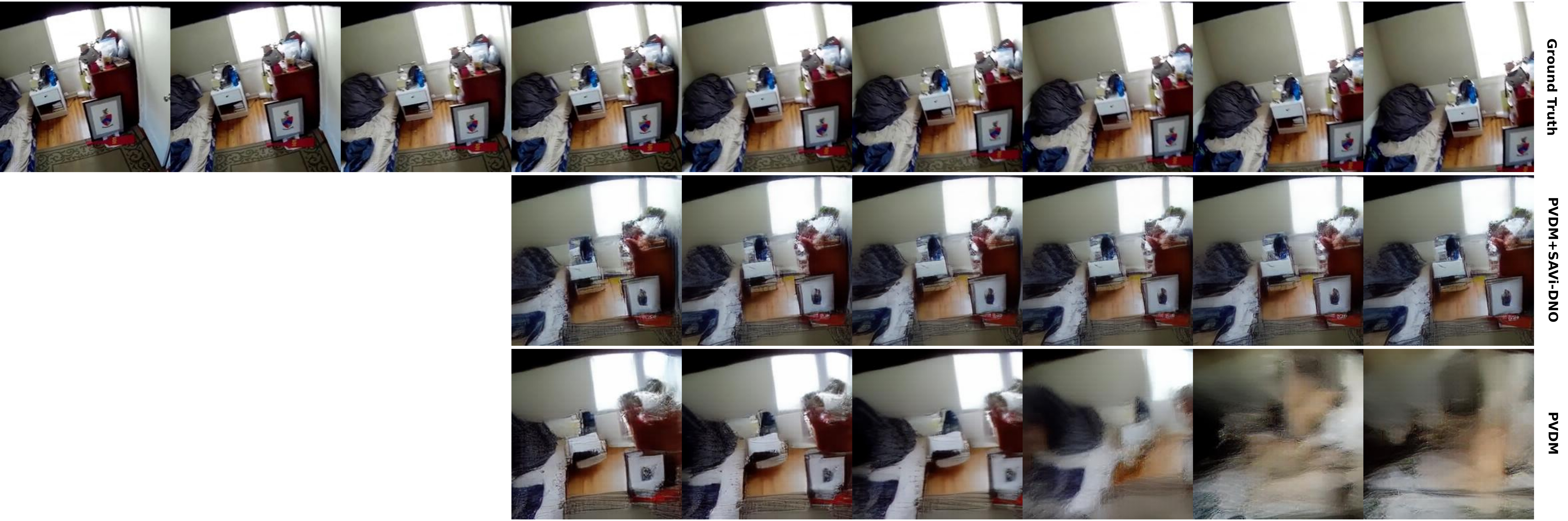}
   \caption{Ego4D qualitative sample.}
   \label{fig:ego4d3}
\end{figure*}

\begin{figure*}[t]
\centering
   \includegraphics[trim=0 0 0 0, width=1.0\linewidth]{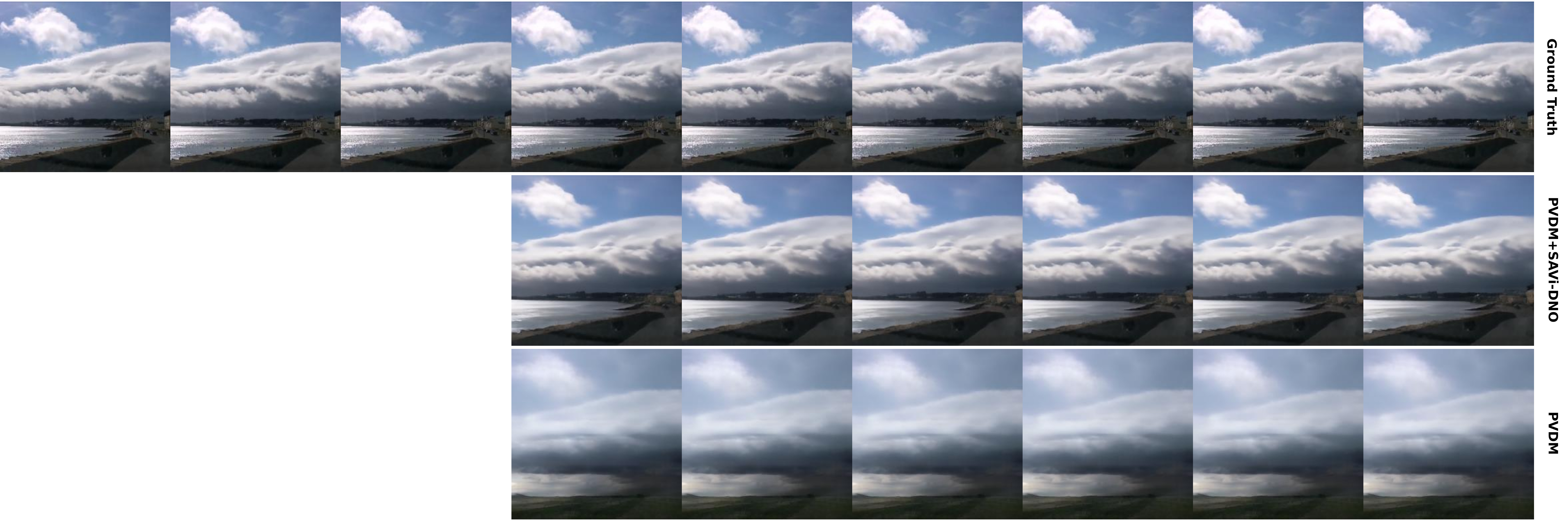}
   \caption{SkyTimelapse qualitative sample.}
   \label{fig:sky1}
\end{figure*}

\begin{figure*}[t]
\centering
   \includegraphics[trim=0 0 0 0, width=1.0\linewidth]{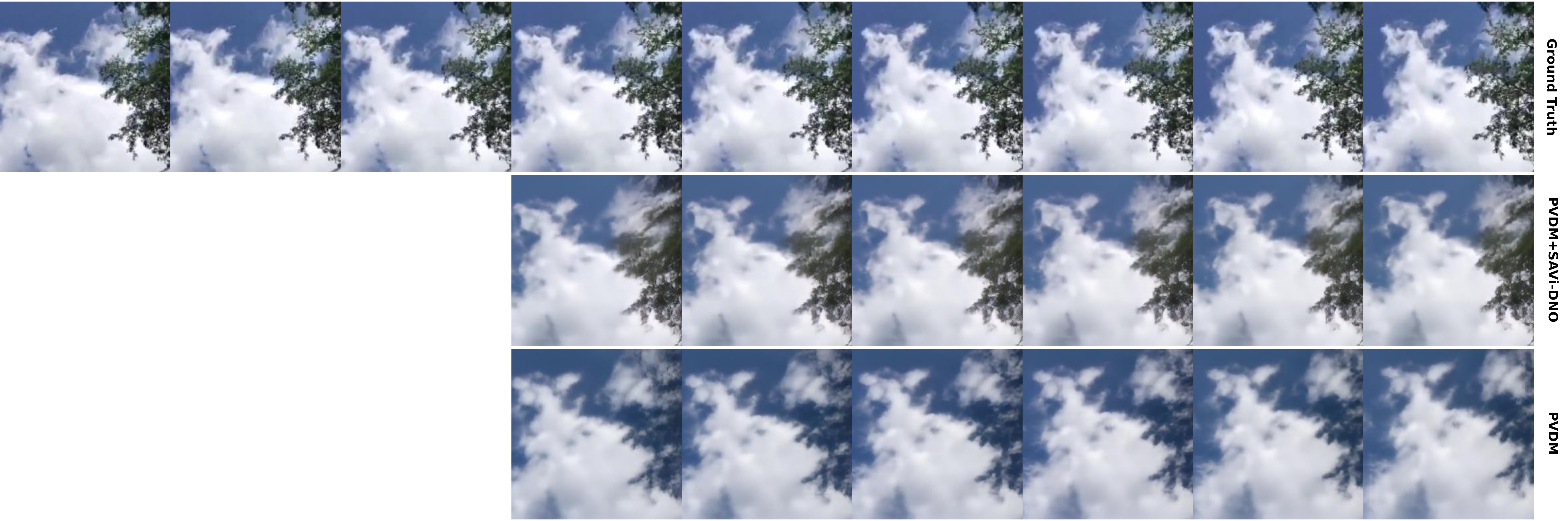}
   \caption{SkyTimelapse qualitative sample.}
   \label{fig:sky2}
\end{figure*}

\begin{figure*}[t]
\centering
   \includegraphics[trim=0 0 0 0, width=1.0\linewidth]{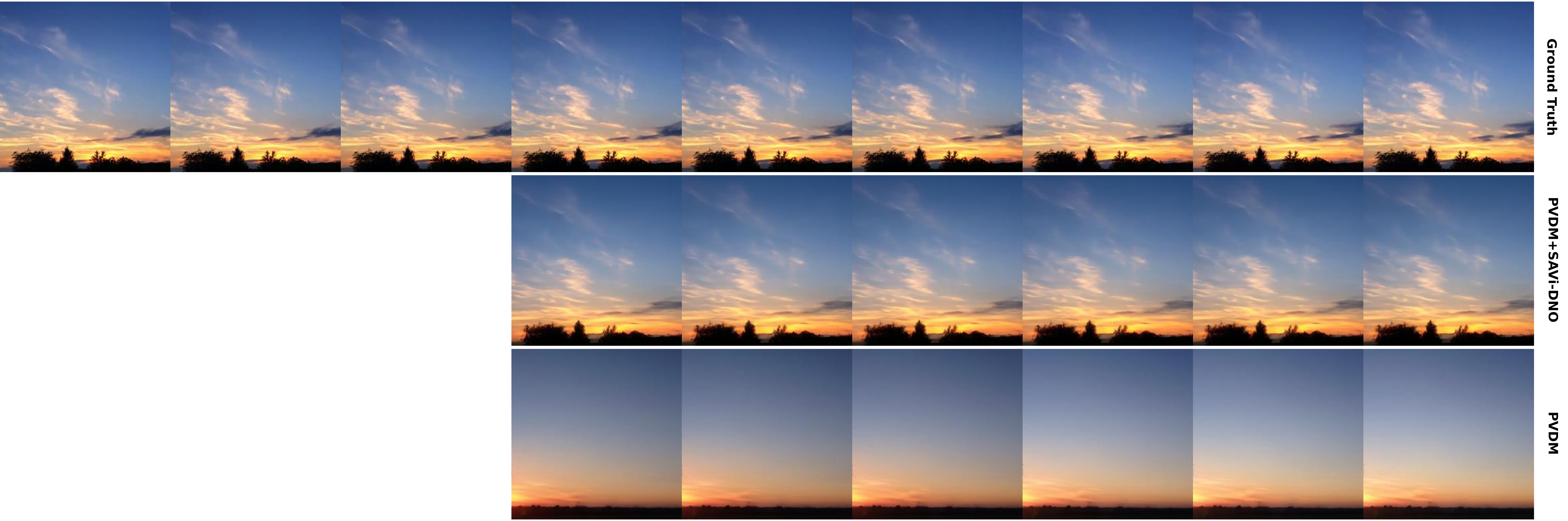}
   \caption{SkyTimelapse qualitative sample.}
   \label{fig:sky3}
\end{figure*}

\begin{figure*}[t]
\centering
   \includegraphics[trim=0 0 0 0, width=1.0\linewidth]{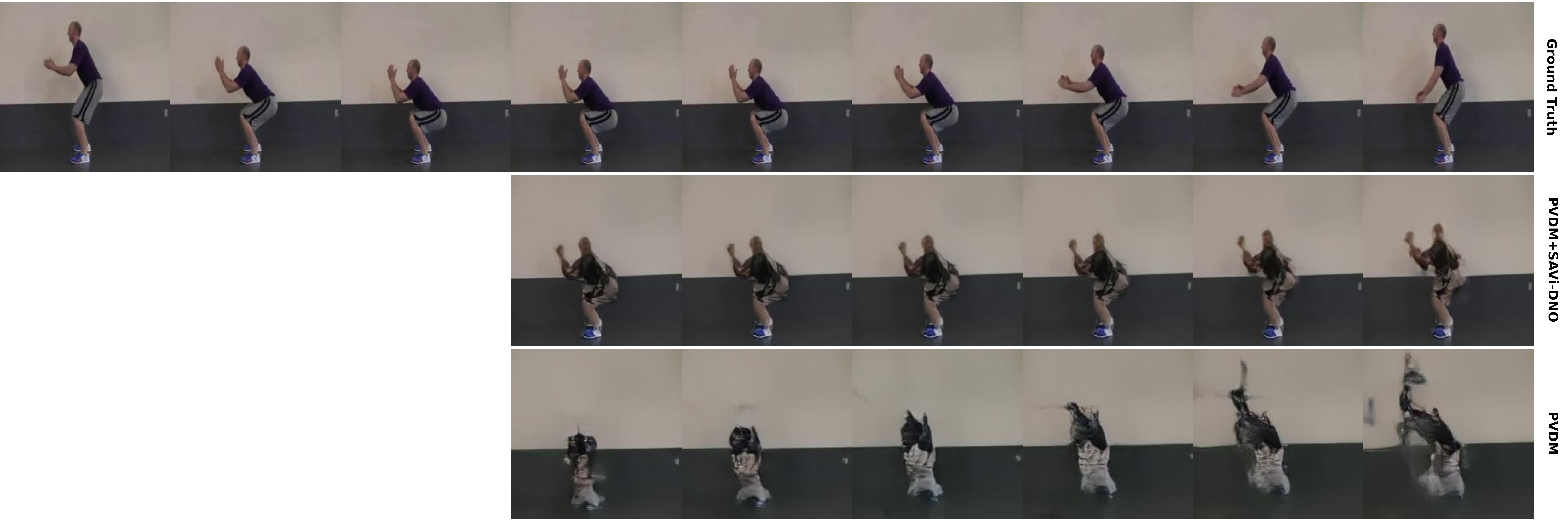}
   \caption{UCF-101 qualitative sample.}
   \label{fig:ucf1}
\end{figure*}

\begin{figure*}[t]
\centering
   \includegraphics[trim=0 0 0 0, width=1.0\linewidth]{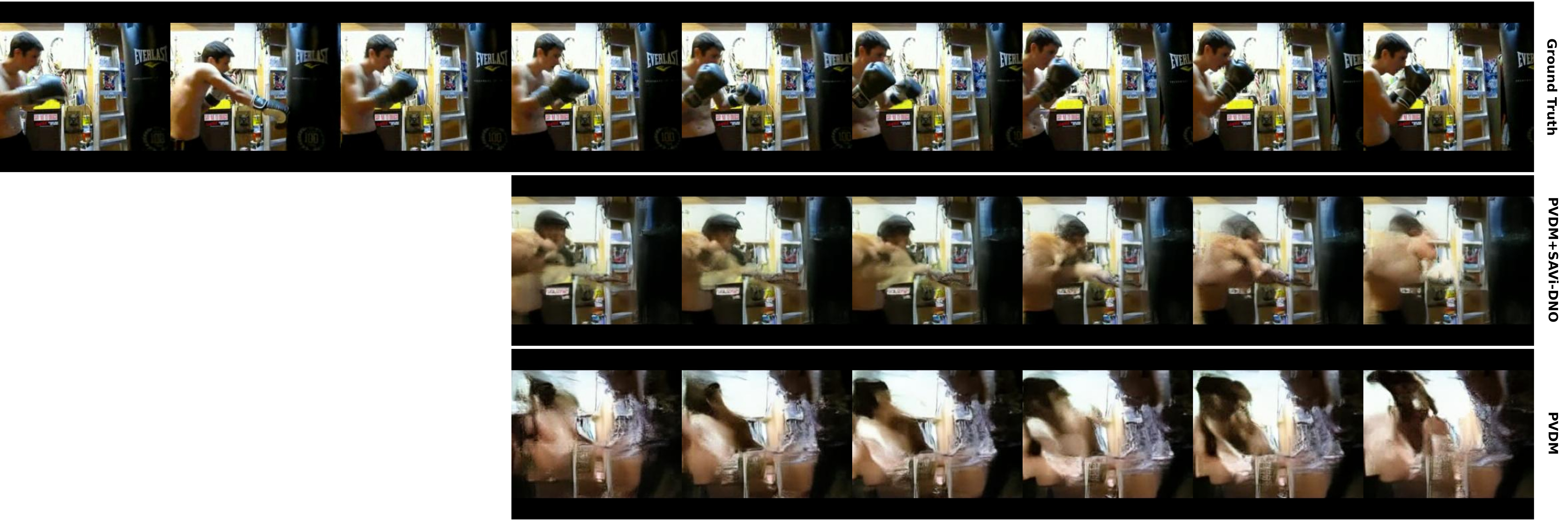}
   \caption{UCF-101 qualitative sample.}
   \label{fig:ucf2}
\end{figure*}

\begin{figure*}[t]
\centering
   \includegraphics[trim=0 0 0 0, width=1.0\linewidth]{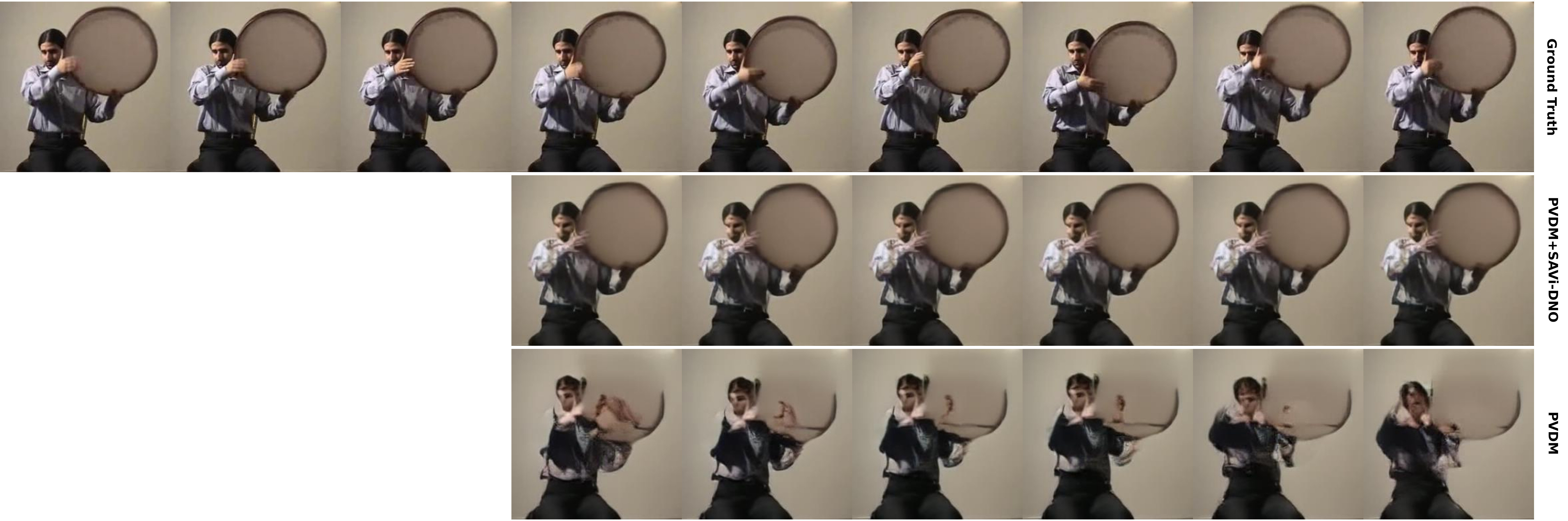}
   \caption{UCF-101 qualitative sample.}
   \label{fig:ucf3}
\end{figure*}

Additional qualitative results on the OpenDV-Youtube, Ego4D, SkyTimelapse, and UCF-101 datasets are provided in (\cref{fig:odv1}, \cref{fig:odv2}, \cref{fig:odv3}), (\cref{fig:ego4d1}, \cref{fig:ego4d2}, \cref{fig:ego4d3}), (\cref{fig:sky1}, \cref{fig:sky2}, \cref{fig:sky3}), and (\cref{fig:ucf1}, \cref{fig:ucf2}, \cref{fig:ucf3}), respectively. Multiple qualitative comparisons also show the benefit of using SAVi-DNO on top of Vista and PVDM.



